\documentclass[runningheads]{llncs}
\usepackage{amssymb} 
\usepackage{amsmath}
\usepackage{indentfirst}
\usepackage{chngpage}

\usepackage{wrapfig}
\usepackage{caption} 
\usepackage[ruled,vlined]{algorithm2e}
\usepackage{graphicx}
\usepackage[utf8]{inputenc} 
\usepackage[T1]{fontenc}    
\usepackage{hyperref}       
\usepackage{url}            
\usepackage{booktabs}       
\usepackage{amsfonts}       
\usepackage{nicefrac}       
\usepackage{microtype}      
\usepackage{bbding}         
\usepackage{hyperref}
\usepackage{xcolor}
\usepackage{tikz}
\definecolor{lime}{HTML}{A6CE39}
\DeclareRobustCommand{\orcidicon}{%
    \begin{tikzpicture}
    \draw[lime, fill=lime] (0,0) 
    circle [radius=0.16] 
    node[white] {{\fontfamily{qag}\selectfont \tiny ID}};
    \draw[white, fill=white] (-0.0625,0.095) 
    circle [radius=0.007];
    \end{tikzpicture}
    \hspace{-2mm}
}

\foreach \x in {A, ..., Z}{%
    \expandafter\xdef\csname orcid\x\endcsname{\noexpand\href{https://orcid.org/\csname orcidauthor\x\endcsname}{\noexpand\orcidicon}}
}
\newcommand{\orcid}[1]{\href{https://orcid.org/#1}{\textcolor[HTML]{A6CE39}{\orcidicon}}}

\begin{document}

\title{\large SMASH: a Semantic-enabled Multi-agent Approach for Self-adaptation of Human-centered IoT}
\titlerunning{SMASH}
%
%
\author{
Hamed~Rahimi\inst{1,2}\orcid{0000-0001-9179-8625} \and
Iago~Felipe~Trentin\orcid{0000-0002-1558-5668}\inst{1,3}\and
\\ Fano~Ramparany\inst{1}\and
Olivier~Boissier\inst{3}\orcid{0000-0002-2956-0533}
}
\authorrunning{H. Rahimi et al.}
%
\institute{
Orange Labs, Meylan, France 
\and
Univ. Lyon, Université Jean Monnet, Saint-Etienne, France
 \and
Mines Saint-Etienne, Univ. Clermont Auvergne, CNRS, \\UMR 6158 LIMOS, Institut Henri Fayol, Saint-Etienne, France\\
\email{\{hamed.rahimi, iagofelipe.trentin\}@orange.com}}
\maketitle              
\begin{abstract}
Nowadays, IoT devices have an enlarging scope of activities spanning from sensing, computing to acting and even more, learning, reasoning and planning. As the number of IoT applications increases, these objects are becoming more and more ubiquitous. Therefore, they need to adapt their functionality in response to the uncertainties of their environment to achieve their goals. In Human-centered IoT, objects and devices have direct interactions with human beings and have access to online contextual information. Self-adaptation of such applications is a crucial subject that needs to be addressed in a way that respects human goals and human values. Hence, IoT applications must be equipped with self-adaptation techniques to manage their run-time uncertainties locally or in cooperation with each other. This paper presents SMASH: a multi-agent approach for self-adaptation of IoT applications in human-centered environments. In this paper, we have considered the Smart Home as the case study of smart environments. SMASH agents are provided with a 4-layer architecture based on the BDI agent model that integrates human values with goal-reasoning, planning, and acting. It also takes advantage of a semantic-enabled platform called Home'In to address interoperability issues among non-identical agents and devices with heterogeneous protocols and data formats. This approach is compared with the literature and is validated by developing a scenario as the proof of concept. The timely responses of SMASH agents show the feasibility of the proposed approach in human-centered environments.
\end{abstract}

\section{Introduction}

The Human-Centered Internet of Things (HCIoT) is a domain of IoT applications that focuses on the accessibility of interactive IoT systems to human beings. Smart Home applications are one of the emerging use cases of the HCIoT that has been directly impacting the lifestyle of people\cite{jiang2004smart}. According to policyAdvice \cite{stasha_depth_2021}, the IoT market will have over \$520 billion revenue in the market by 2027 and Statista \cite{noauthor_smart_2020} estimates Smart Home applications occupy 12.2\% of the market in 2021 that are expected to grow up to 21.4\% by 2025. As the number of Smart Home applications increase, IoT devices are becoming more and more ubiquitous. These objects are responsible for controlling various tasks such as sensing and computing, planning, learning, and acting \cite{muccini2018self}. Hence, IoT objects in Smart Home applications need to use self-adaptation techniques to manage run-time uncertainties in their dynamic environment. One of these techniques is planning and acting  \cite{ghallab2004automated}, which allows intelligent agents, goal-driven entities capable of perceiving and acting upon their environment \cite{russell_artificial_2016}, to plan for their goals and to adapt their behavior at the run-time for achieving them. The integration of planning and acting advocates an agent’s deliberation functions, in which online planning takes place throughout the acting process. In Smart Home applications, where HCIoT consists of several devices, intelligence can be integrated into two levels. The first is in the device layer where we embed intelligence in each device to control its functionality correspond to the behavior of other connected devices. The second level is to implement the intelligence in a CPU that is connected to devices and control their functionality. In this research, we run deliberation functions on multiple intelligent agents with various coordination models at the system level. Therefore, we can describe these applications as a Multi-Agent System (MAS) \cite{wooldridge2009introduction} that is designed to manage such environments with several intelligent agents controlling domestic artifacts of IoT objects. These agents require access to the information of IoT objects for decision-making and handling data exchanges. Semantic Web Technology \cite{berners2001semantic} is an effective way that simplifies information exchange and enhances the interoperability among various devices. 

This paper introduces SMASH: a multi-agent approach for self-adaptation of IoT applications in Smart Home. SMASH agents have a 4-layer agent architecture that autonomously adapts Smart Home devices to uncertainties of their environment by planning and acting of user-centric agents that respect human values and are supported by value-reasoning and goal-reasoning mechanisms. SMASH agents are integrated with a semantic-enabled platform for Smart Home applications called Home’In to create a Multi-agent System capable of autonomous decision-making. The rest of the paper is organized as follows. The state-of-the-art is investigated in Section 2 and the proposed approach is presented in Section 3. In Section 4, we demonstrate the feasibility of the proposed approach by modeling a Smart Home scenario, whose implementation and validation serve as a proof of concept. In Section 5, we compare our results with the related works. Finally, in Section 6 we present our conclusions. 

\section{Literature Review}

HCIoT applications are closely incorporated with humans and their environment and have a lot of benefits for society, the environment, and the economy. A Smart Home \cite{trentin2019insights} is a living environment that is equipped with artificial intelligence that controls the functionality of a house such as heating, lighting, and other devices. Smart Home is one of the HCIoT applications that play an important role in human lives, and the lifestyle of people has a huge dependency on their existence. However, due to direct interaction with human, these application comes with various issues that need to be addressed. Self-Adaptation is one of the issues that focus on functional control elements that interact with the human environment and enabling dynamic adjustment to run-time uncertainties to provide some QoS requirements. Planning and Acting \cite{ghallab2004automated} is one of the strong solutions, in which an agent with proactive behavior, that has planning and deliberation capabilities, is more loosely tied to its environment. By knowing which goals to seek, it can reason upon context and create plans to achieve them. Such reasoning models that integrate planning and acting are explored in \cite{li2020decentralized,jordan2018better,ciortea2018repurposing,meneguzzi2015planning}. In \cite{ghallab2004automated}, an agent is capable of performing planning and deliberative acting. Deliberative acting is defined as an understanding of the surrounding context and intelligently decision-making in a given situation. Deliberative planning is defined as a way to seek the achievement of goals by executing plans, through the realization of tasks, reaching a contextual state, or the maintenance of a condition. In \cite{patra2020integrating,ingrand2017deliberation,de2018htn,xu2018formal,sardina2011bdi}, the challenges of merging planning and acting are addressed. Besides, some approaches are presenting goal-reasoning architectures \cite{niemueller2019goal,pokahr2005jadex} that are practical for adaptation of intelligent agents, which may control IoT objects of the smart environment. Although, these architectures do not consider human values that may directly affect decision-making in different conditions. There are some works on value-reasoning models \cite{vanhee2015using,battaglino2013emotional,boshuijzen2020agent}. However, these works have not explored the integration challenges between planning, acting, goal-reasoning, and value-reasoning functions for human-centric smart environments. These solutions are not commonly self-adapted to human behaviors and uncertainties of the environment, and in some cases perform based on tasks that are once programmed by the administrator. In dynamic environments, where IoT objects have direct interactions with human beings and devices need to have access to the online contextual information, the system requires to adapt its execution to user needs and context evolution of the environment including humans. More specifically, a Smart Home has to be context-aware and adapt its user-centric services at run-time in order to increase human comfort. For instance, it easies answering a call while the user is watching a movie. Ideally, a Smart Home should minimize its direct input from users and be autonomous in the pursuit of user-centric goals to satisfy users’ needs.


\begin{figure}
  \centering
  \includegraphics[width=5.5cm, height=5.5cm]{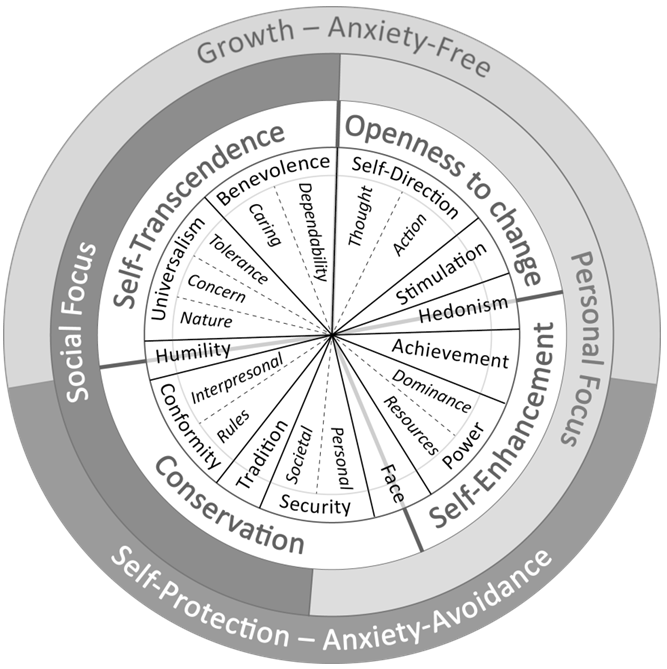}
  \caption{Motivational continuum of human values}
\label{fig:values}
\end{figure}

Human values are an aspect of human subjectivity, which can be described as the criteria used by humans to select and justify actions (guiding human behavior), as well as evaluate people and events (judging behaviors and situations) \cite{rohan2000rose}. Values are stable over time \cite{bardi2009structure}, therefore all value-related configurations of a system need to be performed only once for a long period. They are also intrinsic to each human being, influencing daily activities and long-term decisions. In our approach, which is presented in the next section, we have used the 19 human values defined by Theory of Basic Human Values \cite{schwartz2012overview} that have brought a refinement of the initial theory from \cite{schwartz1992universals}. This theory has been supported by recent cross-national tests \cite{cieciuch2014cross}, and results show that human values can be organized in a continuum (Figure \ref{fig:values}) based on the compatibility of their pursuit. Adjacent values tend to have compatible pursuit, while opposite ones tend to be incompatible as if they have a conflict with each other. This intrinsic characteristic of values is interesting to be considered during reasoning processes since it can help to find priorities and incompatibilities at run time for decision making.

Our approach integrates human values with goal-reasoning, planning and acting of user-centric agents in Smart Home applications. In the next section, we propose a reasoning architecture for intelligent agents that improves user-centricity by applying the new dimension of human values in the goal-reasoning, planning and acting layers. We also explain the integration of these agents with the Home'In platform. The proposed system is able to act autonomously, allowing agents to independently perform actions whilst respecting users’ values.

\section{The Proposed Architecture}
In this section, we address the local adaptation problem by proposing a value-driven agent architecture called SMASH that focuses on the intelligent adaptation of IoT applications in Smart Home and is designed based upon human values. SMASH is a user-centric agent architecture that adapts its reasoning and functioning based on value-driven reasoning and deliberation process consisting of 4 context-aware layers: 1) Value-Reasoning Layer, 2) Goal-Reasoning Layer, 3) Planning Layer, and 4) Acting Layer. The first layer is dedicated to reasoning upon human values, ordering them according to the context of the environment and the user's personal preferences (Section 3.1). The second is dedicated to goal reasoning and identifying goals to be achieved, based on the context and values from the first layer (Section 3.2). The third layer is for run-time planning given the context of the environment,  selected goals, and the given values respectively from the second and first layer (Sec. 3.3). And the last layer is for initiation of acting upon given plans and values, selecting actions to be performed. (Sec. 3.4). This agent architecture supports a high degree of autonomy while respecting the human values of users.
\begin{figure}
  \centering
  \includegraphics[width=0.6\linewidth]{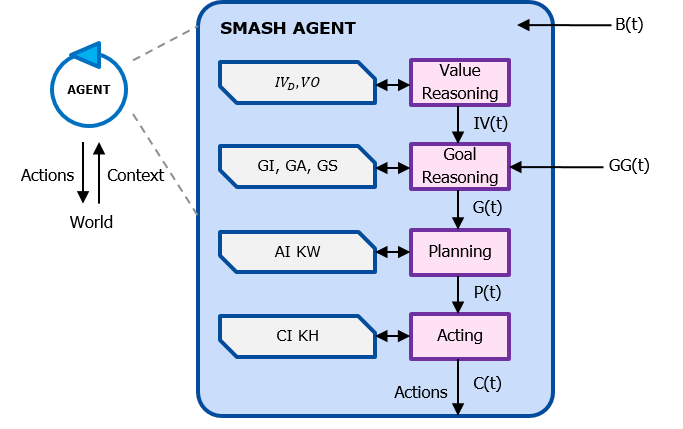}
  \caption{SMASH Agent Architecture Overview}
\end{figure}

\subsection{Value-Reasoning Layer}

This layer is provided with a set of ordered values $IV_D(t)$, which shows the Default Importance of Values and indicates the user’s priorities and preferences toward these human values at time $t$. This set of values alongside with a set of value ordering rules $(VO)$  are able to autonomously reorder the values based on the context of the environment and the preferences of the user. This layer provides the next layers with a set of totally ordered values $IV(t)$. In (1), $B(t)$ is the set of beliefs that represents the context of the system at time $t$.
\begin{equation}
    IV(t)=ValueReasoning~(B(t), IV_D, VO)
\end{equation}
\paragraph{Value-ordering rules}
Each rule of $VO$ has the format $``condition \rightarrow body"$, where $condition$ is a first-order logical formula stating if the rule is active in the current context. The $body$ is a formula with a binary or unary operator that aims to sort the values as follows: 
\begin{itemize}
  \item The operator $(\ll v)$ and $(\gg v)$ is responsible to set the value $v$ respectively as the least and most important value in the set $IV(t)$.
  \item The operator $(\succ v_1\ v_2)$, $(\prec v_1\ v_2)$, and $(\sim v_1\ v_2)$ makes the value $v_1$ respectively greater, lesser, and same as the importance of value of $v_2$ in the set $IV(t)$.
   \item The operator $(-\ v)$ deletes the value $v$ from the set $IV(t)$.
\end{itemize}
It must be noted that $IV_D$ may, at a certain moment, not contain all 19 values since the user may consider some values as not important by default. Besides, the order of rules defined in $VO$ is important, since the execution of the formulas in the $body$ is done following that order. For instance, if $VO = \{ (c_1 \rightarrow (\gg v_1) ), (c_2 \rightarrow (\succ v_2\ v_1) ) \}$, in case both $c_1$ and $c_2$ is evaluated as $True$ conditions, firstly, the value $v_1$ is put in the first position (i.e. the most important value), and by applying the second rule, the value $v_2$ takes its position and becomes more important than the value $v_1$ in $IV(t)$. 
\subsection{Goal-Reasoning Layer}

\begin{figure}[t]
\centering
\includegraphics[width=0.8\linewidth]{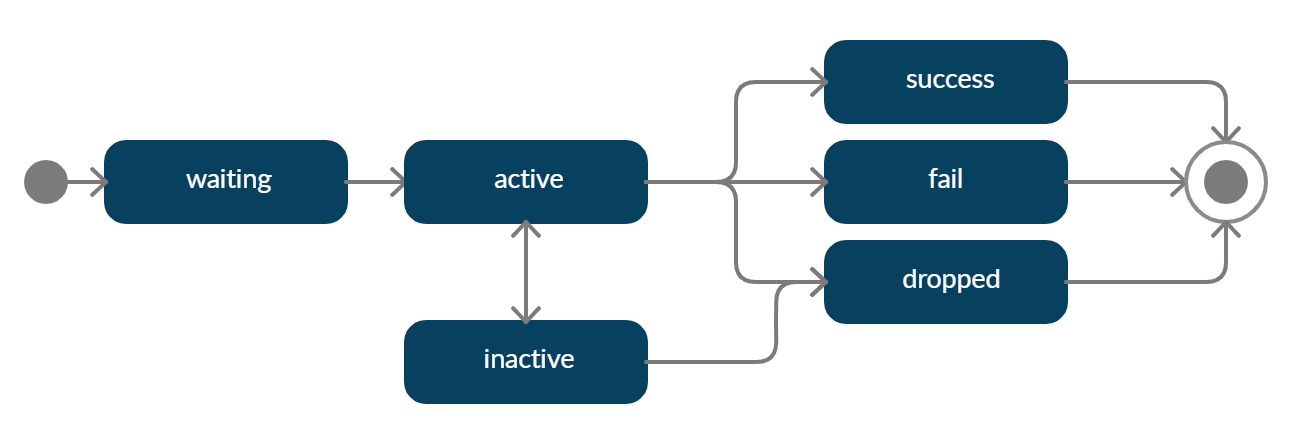}
\caption{State diagram of goal states}
\label{fig:goal-states}
\end{figure}

This layer manages and activates a set of goals $G(t)$ from the belief base $B(t)$ and the set of ordered values $IV(t)$ that is the output of the previous layer. The goal-reasoning performs on the goals given by the user $GG(t)$ and the set of existing goals called Goal Status $GS(t)$. These given goals along with the context of environment $B(t)$ update and filter $GS(t)$ based on activation rules $GA$ and goal impact rules $GI$. Note that the impact of activated goals on the user’s values is also based on the given values $IV(t)$ and the current context $B(t)$.

\begin{equation}
G(t)= GoalReasoning(B(t), GG(t), IV(t), GS(t), GA, GI)
\end{equation}
As shown in equation (2),  $G(t)$ is an ordered subset of goal statuses $GS(t)$ that contains newly created or updated goals at time $t$. Besides, GS(t), subset of B(t), is a set of beliefs of the form $state(goal, status, source)$ that describes the $status$ of a $goal$ that is from a $source$. $source \in \{user, self\}$ indicating the source of the goal that could be directly given by the user or could be autonomously activated by the reasoning of the agent. The goal-activation rules from $GA$ are able to change the status of goals. As shown in Figure \ref{fig:goal-states}, the $status \in \{waiting, active, inactive, success, fail, dropped\}$ that states as follows:
\begin{itemize}
  \item The status $active$ is for goals that are activated to be pursued by the agent. 
 \item The status $waiting$ is for the goals that are on hold to be processed. 
 \item The status $inactive$ is for the goals that are disabled to be pursued by the agent. 
 \item The status $fail$ is for the goals that were not successful to be achieved.
 \item The status $success$ is for the goals that were achieved successfully.
 \item The status $dropped$ is for the goals that are no longer desired to be pursued.
 \end{itemize}

\paragraph{Goal-activation rules}
$GA$ rules have the format $(condition \rightarrow body)$, where $condition$ is a first-order logical formula, that checks the state of the environment through $B(t)$. The $body$ is a list of first-order propositions with the format $state(goal, status, source)$ that updates the status of the goal issued of the source in $GS(t)$. For instance, as shown in equation (3) there is the default rule in $GA$, which is responsible to activate the waiting goals given by the user, known as $GG(t)$. 

\begin{equation}
    state(Goal, waiting, user) \rightarrow state(Goal, active, user)
\end{equation}

\paragraph{Goal-impact rules}
The $GI$ rules are responsible to define the impact that a goal with a certain status, under a certain contextual condition, has over a value. These rules check if goals are respecting the values.
The format of these rules is expressed as $(condition \rightarrow body)$, where $condition$ is a first-order logical formula that checks the context of the environment, and $body$ is a tuple of $(goal, impact, v)$ that checks the positive, neutral or negative $impact$ of a $goal$ on the value  $v$.

\begin{figure}[t]
\centering
\includegraphics[width=.7\linewidth]{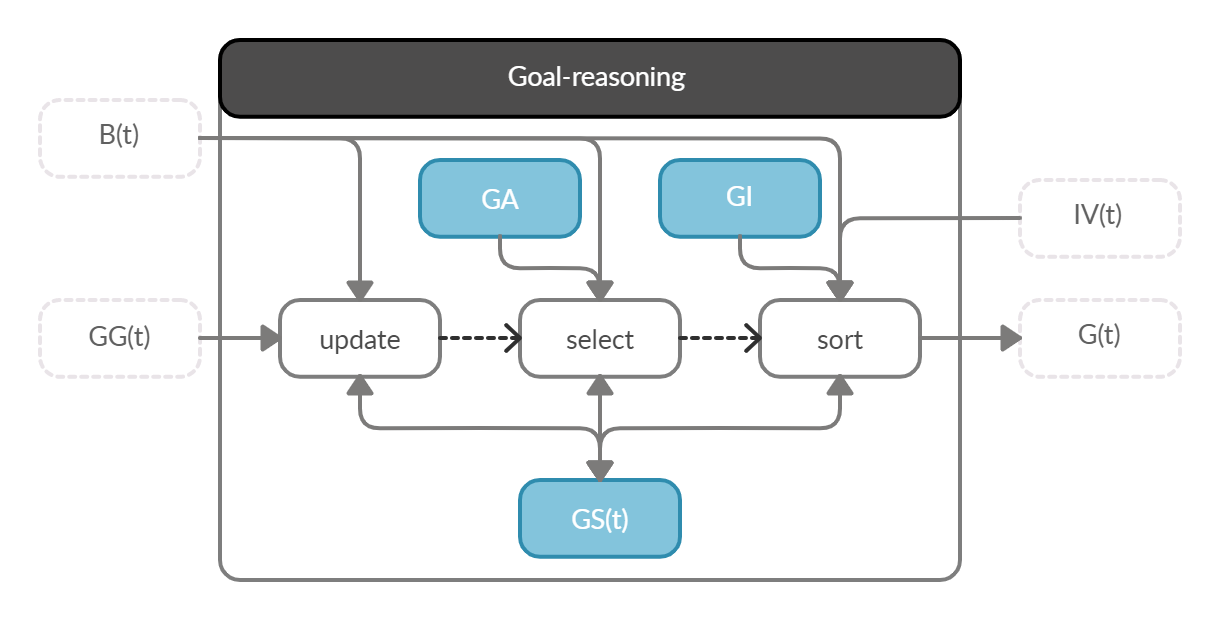}
\caption{Visual representation of goal-reasoning layer}
\label{fig:goal-algo}
\end{figure}

Figure \ref{fig:goal-algo} presents a visual representation of the goal reasoning layer. The elements $update$, $select$ and $sort$ are the functions executed during one reasoning cycle. 

\begin{itemize}
\item The function $update$ adds all goals in $GG(t)$ with the status $waiting$ to $GS(t)$, in case their $source$ is equal to $user$. Note that the status of goals in $GS(t)$ also gets updated in the Planning Layer and Acting Layer with $success$ or $fail$.

\item The function $select$ applies all goal-activation rules from $GA$ whose condition is satisfied in $B(t)$, i.e. the $body$ of these rules are used to update $GS(t)$ (setting statuses to $active$, $inactive$, or $dropped$). These rules allow the activation of goals based upon context, and these new goal statuses have $self$ as their $source$. If necessary, goal activation rules can also be created to prevent the duplication of goals, by dropping any new goal that already exists in $GS(t)$.

\item The function $sort$ filters all the elements in $GS$, which represent the goals and their statuses; the only goals sorted are those whose $source$ is equal to $self$, and the sorting is based upon the $GI$ rules that are initiated in the current context. The goal that have negative impacts over an existing value in $IV(t)$ are deleted from $GS(t)$ by this function.
The importance order of values $IV(t)$ is used to sort $G(t)$ according to the goal impacts of $GI$.
\end{itemize}

\subsection{Planning layer}
\label{sec:pl}

The Planning Layer generates a set of plans in real-time to achieve goals of $G(t)$ provided by the previous layer. The set of plans $P(t)$ is created based upon the context $B(t)$, the ordered set of values $IV(t)$, the ordered set of goals $G(t)$, the Action Impact rules $AI$, and a set of tuples Know-What (KW), which shapes the action model depending on the context by providing a set of right actions of the system. The action-impact rules $AI$ describe the impact of actions over the values of $IV$. A plan $(goal, body)$ in $P(t)$ is a tuple, where $body$ is the list of actions that need to be executed by the acting layer in order to achieve the $goal$.

\[
P(t) = planning(B(t), IV(t), G(t), KW, AI)
\]

\paragraph{Know-what}
The elements of $KW$ are tuples of the form $KW = \{ kw_i\ /\ kw_i = (action, condition, effect) \}$. They express the action models, where $kw_i.action$ describes the name of action and its parameters, the $kw_i.condition$ is a first-order logical formula with contextual condition under which the action can be executed, and $kw_i.effect$ includes the tuple $(add, delete)$ that define the action $add$ or $delete$ on agent's beliefs after the execution of the plan. 

\paragraph{Action-Impact rules}
The Action-Impact $AI$ represents the impact of an action, under a certain contextual condition, over a value. They are defined as $(condition \rightarrow body)$, where $condition$ is a first-order logical formula based on the current context, and $body$ is the tuple $(action, impact, v)$, presenting the positive, neutral or negative $impact$ of the action $action$ over the value $v$. All the action-impact rules $AI$ are created based on the Command-Impact rules ($CI$) that is explained in next subsection. Every $kw_i.action$ has a $kh_j.body$ in the tuples $Know-how (KH)$ describing the positive, negative, or neutral impact over value $v$.

The Fast Downward (FD) planner \cite{helmert2006fast} is a configurable forward state-space search planner that is responsible to perform the planning in this layer. The planning layer translates the necessary information into Planning Domain Definition Language (PDDL), giving to the planner two inputs: 
\begin{itemize}
\item a domain definition PDDL file, where elements from $B(t)$ are added as $predicates$, and elements from $KW$ are translated as $actions$;
\item a problem definition PDDL file, where $B(t)$ are translated into propositions representing the $initial\ state$, and goal statuses from $G(t)$ are translated into individual $goals$.
\end{itemize}

Before translating $KW$ elements into $actions$, the set $AI$ is used to filter all actions that have a negative impact over values presented in $IV(t)$ in order to respect the values of the user. The planner is responsible to compute a plan to every goal in $G(t)$ and pass it to the acting layer for execution. If planner doesn't succeed in finding a plan for the goal, it will then update its goal state in $GS(t)$ with the status $fail$.

\subsection{Acting layer}
\label{sec:al}

Acting layer receives a set of plans $P(t)$ from the previous layer and is responsible to initiate and execute a set of actions aiming to achieve the given goals. The execution of a plan consists of the constant refinement of selected action until it obtains a list of commands $C(t)$ to be executed at time $t$. For this refinement, the acting layer requires a set of tuples Know-How ($KH$), which are the execution model of actions according to the contextual condition of the system, and Command Impact rules CI, which describe the impact of selected commands over the user’s values.

\[
C(t) = acting(B(t), IV(t), P(t), KH, CI)
\]

\paragraph{Know-how}
The tuples $KH = \{ kh_j\ /\ kh_j = (action, condition, body) \}$ are the execution bodies of actions, where $kh_j.action$ is the action name and its parameters, the $kh_j.condition$ is the first-order logical formula with contextual condition under which the action can be executed, and $kh_j.body$ is an ordered set of commands to be executed, and actions to be refined into commands. For every action in $KH$ there is an action model in $KW$, in which $kh_j.action = kw_i.action$.

\paragraph{Command-impact rules}
The $CI$ defines the impact of command, under a certain contextual condition, over a value. They are expressed as $(condition \rightarrow body)$, where $condition$ is a first-order logical formula bearing on the current context, and $body$ is the tuple $(command, impact, v)$, presenting the $command$ being considered, and positive, neutral or negative impact over the value $v$.

\section{Integration between SMASH agents and IoT Layer}
Smart Home applications are increasingly becoming larger and more complex due to the spread of various ubiquitous appliances that are supported by different vendors. In addition to the increase in the number of devices, the stack of messages, data formats, communication protocols, and system architectures are even so evolving.
Facing such heterogeneity, several standardizations such as ETSI SmartM2M SAREF \cite{lefranccois2017planned} and W3C Web of Things (WoT) \cite{kovatsch_web_2019} try to address the syntactic and semantic languages of these systems. Existing solutions to integrate different IoT services are mainly based on manual IoT mashups (e.g. NodeRed) \cite{guinard2009towards}, which are a way to compose a new service from existing ones \cite{kovatsch2015practical} in order to allow the synergy among heterogeneous devices, services, protocols, and formats. Even in ad-hoc mashups that are often temporary, heterogeneity is limited, and services are provided for specific needs \cite{guinard2009towards}. Some solutions discover devices and their associated services using existing Web infrastructure. Although, these solutions face several issues such as connection and integration. Semantic Web Technologies (SWT) \cite{berners2001semantic} are a high-level contextual language that abstracts communications and creates a common request format for accessing different services on the web infrastructure. Semantic-enabled IoT systems are an effective solution that addresses the complexity of context management \cite{euzenat2008dynamic,perera2013context} and improves interoperability among heterogeneous devices. This interoperability allows the system to have access to various contextual information, which helps to take advantage of various adaptation techniques that require more contextual resources. 

This section presents the integration between SMASH agents and the Home’In platform, which is able to interact with various devices in Smart Homes and address their interoperability issues. SMASH manages to autonomously adapt the Smart Home devices to uncertainties of their environment respecting the user’s values by reasoning based on planning and acting. Fig. 1 shows the proposed approach which is consisting of a multi-agent system and Home’In platform that is connected to real devices. The approach firstly represents a multi-agent system composed of SMASH autonomous agents that are responsible to satisfy users’ goals considering their values in real-time. Following a multi-agent oriented programming approach \cite{boissier2013multi}, the agents have uniform access to the resources and tools of the environment via a set of non-autonomous entities called artifacts. In Smart Home applications, artifacts are responsible to encapsulate the access and control of smart devices such as TV, Phone, or Sofa.  In Home’in, the Context Manager receives information from the physical world via devices connected to Home’In. These devices frequently send data to dedicated Home’In service providers that are responsible to consider different abstraction layers with various formats and content, and then store them in the Context Manager as RDF triples to update the system in real-time. In order to provide context for the agent, we created an interface with an already existing industrial solution working as a context manager. This industrial solution provides two channels for retrieval and storage of contextual information:
\begin{itemize}
\item  a REST API with an HTTP endpoint for context retrieval, that receives SPARQL requests and replies with JSON or XML responses (both using predefined ontologies selected by the industry responsible for the solution); and
\item an MQTT broker for context storage, that receives and replies JSON messages coded in proprietary format.
\end{itemize}


Our contextual information is accessed through request/response (e.g. REST) or publish/subscribe (e.g. MQTT) patterns. The IoT Devices Manager provides a direct interface to contact the devices that are located in the physical environment of the home and are connected to Home’In. In other words,  It sends commands to these devices in order to make them perform actions or to retrieve the information they’ve sensed in the environment. Home’In components are interconnected through MQTT and Rest buses that enable the Context Manager to listen to abstraction layers data and store selected ones through IoT Devices Manager for representation of the context. The SMASH architecture is implemented via Multi-Agent Oriented Programming (MAOP) \cite{boissier2013multi,boissier2019dimensions} that keeps the design modular and distributes the tasks and responsibilities between SMASH agents. The heterogeneity of devices is not the concern of agents due to providing a uniform set of actions and observable properties into artifacts connected to Home’In. Artifacts provide information to agents through observable properties, and access to actuators through operations. The description of these features defines what we call the user interface of the artifacts. They also produce signals in case of success/failure of operations or transient information produced during execution.

\begin{figure}[t]
\centering
\includegraphics[width=.5\linewidth]{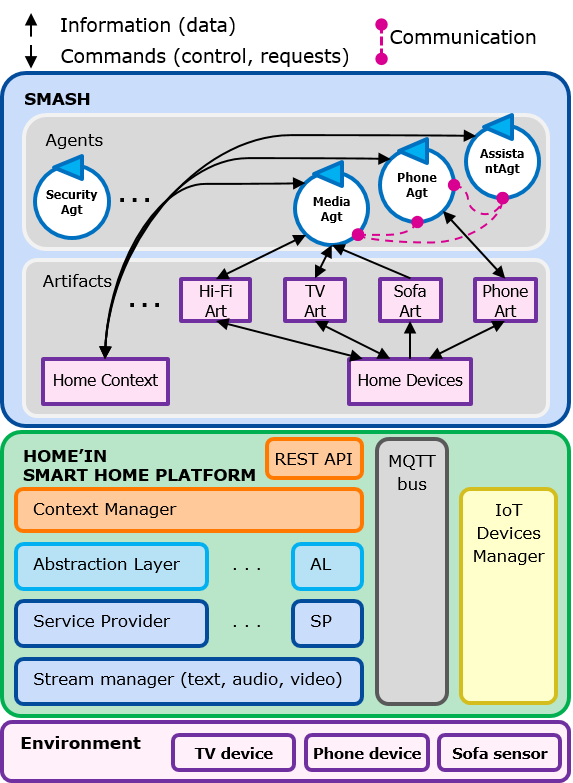}
\caption{Integration of SMASH Agents and Home'In}
\end{figure}

\section{Proof of Concept}

\begin{figure}[h]
\centering
\includegraphics[width=.6\linewidth]{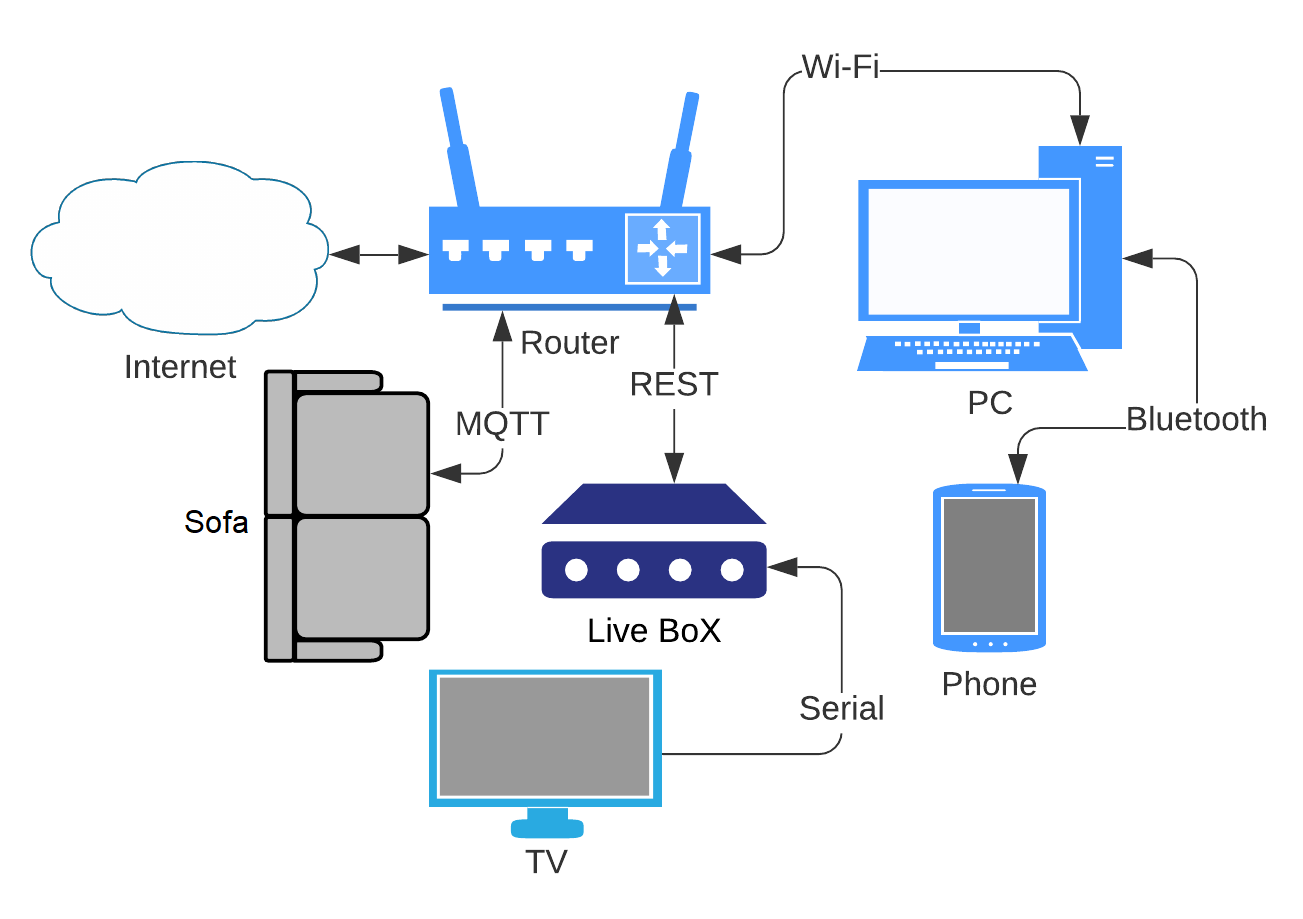}
\caption{Overview of Technical Infrastructure}
\label{fig:OoI}
\end{figure}

In this section, we aim to design a scenario in Smart Home applications and implement the proposed approach to address self-adaptation of IoT devices. The scenario and the SMASH agents are implemented in a multi-agent oriented programming platform \cite{boissier2020multi} called JaCaMo. The acting layer uses the built-in BDI engine as previously explained, the planning layer uses the PDDL language and the FD planner \cite{helmert2006fast} as described in Section \ref{sec:pl}, and the goal- and value-reasoning layers are implemented in Java and are available to agents as internal actions in JaCaMo. An internal action allows JaCaMo to call Java functions defined by the developer. As shown in Fig. \ref{fig:OoI}, There are a PC, a Smart Phone, a Smart TV, and a Smart Sofa in the smart environment of the scenario.

\subsection{Scenario}
In this short scenario, we have a scene where Max, the user, asks the Intelligent system of the house through his phone to play a program on TV. Assuming TV is off, the intelligent systems initially turns TV on and once Max arrives in the living room and sits on the smart sofa, which identifies him, TV starts playing the program. After a few minutes, Max receives a phone call from his boss. Considering Max is out of duty, the pleasure of continuing watching TV (that is represented by the value Hedonism in Theory of Human Values) is more important than answering the the working call (that is represented by the value Conformity-Rules in Theory of Human Values). Therefore, the system automatically puts his call on the voice mail. A few minutes later, Max receives a call from his mother. Respecting Max's value, in which the value Benevolence-Caring that represents family relations is more important than Hedonism in his value base, the phone starts ringing and the TV goes on mute. The sofa detects the absence of Max due to his leave toward answering the phone. Max, after responding to the phone call, comes back to the living room and sits on the sofa. Right away, the television unmutes itself and continues playing the channel.

\subsection{Technical Overview}
Using the above scenario, we will go through the reasoning process of SMASH agents and show the transition of states and the exchange of messages. Max is in his leisure time at home, therefore, value reasoning layer compute the order of the values correspond to his current contextual information. In his belief base, the value 'Benevolence-Caring' noted vbc is the most important value, and the value 'Hedonism' noted vhe is more important that the value 'Conformity-Rules' noted vcr.

 \[
 V_{vbc}~>~ V_{vhe}~>~V_{vcr}
\]

Max sends a request to SMASH through his phone that is connected to Home'In platform for watching ``Canal+''. This request is translated as a goal in the set GG(t) of the system. In result, a JSON message is sent to the IoT Device Manager of Home'In over MQTT and Goal reseanong layer starts its computation. The messages exchanged are as follows:
 \[
  Goal~~to~~achieve:~~watch(TV,Canal+) 
\]
IoT device manager shares this message with agents to achieve the goal $watch(TV,Canal+)$. Considering that TV is off, the goal activation rules activate another goal to turn the TV on and save another goal for broadcasting the program once the user sits on the sofa.
 \[
  TV~~is~~OFF  ~ \implies ~ goalActivation(turnOn(TV), active,self)  
\]
The planning layer find the right functions to achieve the goals received from the goal reasoning later. Then, the Acting layer, turns the TV on and put it on the Standby Status to save energy. 
 \[
  +deviceStatus(TV,Standby)
\]
Sofa is intelligent and can identify the user through his weight using a classification method with supervised learning. Once the Max sits on Sofa, the goal reasoning layer and planning are activated, and then by acting layer, the TV starts displaying the ``Canal+''. 

 \[
  +beSeated(Max,Sofa)~ \implies ~deviceStatus(TV,Playing)
\]
Throughout watching TV, Max receives a phone call from his boss. So, it adds a new belief to the system. 
 \[
  +callerType(Boss,Work)
\]
Considering the user's value at the moment, in which pleasure (vhe) is more important than work hierarchy subordination (vcr), the goal reasoning layer doesn't notify Max and activates a goal to set the phone call on Voicemail. 
 \[
  V_{vhe}~>~V_{vcr} ~ \implies ~~  goalActivation(Voicemail(Phone), active,self)
\]
Therefore, the acting layer puts the call directly on Voicemail.
 \[
 +deviceStatus(Phone,Voicemail)
\]
Later, when Max receives a call from his mother, the goal reasoning layer activate a goal for muting the TV and notifying the user by putting his phone on the Ringing Status based on his value at the moment, which is family (vbc) is important than pleasure (vhe). 
 \[
  +callerType(Mom,Family);
\]
 \[
  V_{vhe}~<~V_{vbc} ~\implies ~ goalActivation(notifyUser(Phone,User), active,self)
\]
In result, the planning and acting layer find a sequence of actions to satisfy the goals.
 \[
  +deviceStatus(TV,Mute)
\]
 \[
+deviceStatus(Phone,Ringing)
 \]
Once the user gets up from Sofa to go to his room to answer the call, Sofa detects his absence and the goal activation rules activates a goal to record the program that Max was watching.
\[
+isStand(User)~\implies ~ goalActivation(recording(TV,Program), active,self)
 \]
 
After a while when he comes back and sits on the Sofa, the process of reasoning is repeated and in result, TV umutes and user can resume watching the program.
\[
+beSeated(Max,Sofa)~\implies ~ goalActivation(resume(TV,Program), active,self)
 \]
 \[
 +deviceStatus(TV,Playing);
 \]

\subsection{Discussion and Validation}
\label{sec:discussion}

The main objective of the paper is to push forward the research interest and development of smart environments, specially the intelligent home, a residential space that takes into account more subjective user aspects such as human values, creating an intelligent autonomous home that better understands its users. The BDI (Belief-Desire-Intention) agent has a basic reasoning function that can be mapped to the deliberative acting function described in the paper. The plan refinement in a BDI architecture is performed by using the agent's procedural plans, which are action models i.e. methods with contextual conditions, to execute plan steps according to the context. It means that a plan can be adapted at run-time thanks to the acting function. Our approach presents the SMASH intelligent agent architecture, which is composed of context-aware reasoning functions. Our result represents a functional proof-of-concept for a multi-agent approach for a Smart Home. 

In the Table \ref{tab:exec-time}, we have shown the reasoning time of the execution of an agent for two different planning cycle. The execution is repeated four times, in order to show the consistency of the reasoning time of the agent for the same reasoning cycle. For instances, in the second execution, the total reasoning time is equal to 2.008 seconds, in which 0.844 s is for the planning of the goal A that is the request of Max for watching ``Canal+'', 0.834 s for the planning of the goal B that is turning on the TV, and 0.330 s for the value- and goal-reasoning of the agent. The performance of the agent, with the additional reasoning steps, showed no significant processing overhead in all studied scenarios. This proof-of-concept validates the  feasibility of the proposed approach and shows the agent usability and timely responses in human-centered smart environments.

\begin{table}[h]
\begin{adjustwidth}{-.5in}{-.5in}  
\begin{center}
\begin{tabular}{c|c|c|c|c}

\multicolumn{1}{c|}{\textbf{Time (s)}}                                                             & \textbf{Execution \#1} & \textbf{Execution \#2} & \textbf{Execution \#3} & \textbf{Execution \#4} \\ \hline
Planning Time for Goal A                                                                            & 0.828                  & 0.844                  & 0.808                  & 0.950                  \\ \hline
Planning Time for Goal B                                                                            & 0.837                  & 0.834                  & 0.791                  & 1.124                  \\ \hline
\begin{tabular}[c]{@{}l@{}}Value- and Goal-Reasoning Time\\ \end{tabular} & 0.425                  & 0.330                  & 0.339                  & 0.408                  \\ \hline
\textbf{Total Time (s)}                                                                             & \textbf{2.090}         & \textbf{2.008}         & \textbf{1.938}         & \textbf{2.482}     
\end{tabular}
\caption{Reasoning Time of an agent for two planning cycles}
\label{tab:exec-time}
\end{center}
    \end{adjustwidth}
\end{table}
\section{Related Work}
\label{sec:related}

The aforementioned work presents a new approach to improve the reasoning process of an intelligent agent designed for smart environments. Such reasoning process includes planning and acting deliberative functions. The proposed approach presents a new reasoning method to improve the self-adaptation process of a multi-agent system for the management of IoT objects in human-centered environments such as Smart Home. Such reasoning process includes planning and acting deliberative functions. This section elaborates the main similarities and differences among related literature, specially concerning the goal-reasoning, planning and acting, whether or not they present a value-driven approach.
An extensive definition and theoretical models for planning and acting are presented in \cite{ghallab2004automated}. Also, in \cite{li2020decentralized} authors proposed a decentralized multi-agent planning and acting, and in \cite{torreno2017cooperative} Torreno et al. surveyed about cooperative multi-agent planning. In \cite{ciortea2018repurposing}, Ciortea et al. designed a Multi-agent System for the Web of Things using planning and acting, and in \cite{meneguzzi2015planning} authors surveyed about the integration of planning algorithms and agent. In all these works, the goal- and value-reasoning functions are not present. The planning and acting functions are implemented using different technologies and platforms.  In our architecture the acting function is matched with built-in elements present in the BDI agent model \cite{rao1996agentspeak}, where for instance \textit{refinement methods} in acting are \textit{plans} in BDI. And our planning function translates to the known PDDL standard for execution in the FD planner \cite{helmert2006fast}. The used planning language contains \textit{domain actions} that matches to the \textit{know-what} of the proposed agent architecture, and the \textit{problem initial state} is built upon the set of \textit{beliefs} present in our agent. Furthermore, there are works illustrating how flexible the planning and acting functions are, covering integrations with hierarchical operational models \cite{patra2020integrating},
with hierarchical task networks (HTN) \cite{de2018htn},
with first-principles planning (FPP) \cite{xu2018formal},
with the belief-desire-intention (BDI) agent model \cite{sardina2011bdi},
and with other deliberation functions such as monitoring, observing, and learning \cite{ingrand2017deliberation}.

The lack of a goal-reasoning function in the BDI model was already identified in previous research \cite{pokahr2005jadex}, which proposes a goal deliberation strategy to mainly avoid conflict in the pursuit of goals. A goal reasoning model is proposed in \cite{niemueller2019goal} that aims at enabling reasoning about the organization and representation of the program flow of planning and execution.
The main difference with the SMASH agent architecture is the value-reasoning function and the consideration of human values, meanwhile the common aspects are the explicit representation of goals, the existence of a goal lifecycle, and formalisms to model the transition of a goal through its lifecycle's states. Propositions of value-driven designs for intelligent agents are increasingly studied in the literature.
Official organizations and governments initiatives such as IEEE Ethics In Action \cite{ieee_ethics_2020} and AI HLEG \cite{hleg_high-level_2020} aim at standardizing approaches for the inclusion of values in system designs, as well as understand and inform about the implications of such subjective elements in the reasoning of agents.
Previous works explored the integration of values in the agent reasoning: \cite{vanhee2015using} aimed at supporting coordination through values; and \cite{battaglino2013emotional} treated values as goals that are activated when context reveals they are at stake, whereas in our architecture values are first-class citizens and are used in every reasoning function and processes. A next possible step is adding a DQN (Deep Q-Network) agent that has learning capability via Deep Reinforcement Learning to the proposed multi-agent system. This agent, being capable of observing the behaviour of the user, may be able to make more precise decisions.

\section{Conclusion}
In this paper, we presented a semantic-enabled approach for self-adaptation of IoT objects in human-centered smart environments. This approach manages to autonomously adapt the ubiquitous devices to the uncertainties of the environment by performing a reasoning based on planning and acting that respects the values of the user. The proposed approach takes advantage of Semantic Web technologies using a platform called Home'In, which addresses interoperability issues and makes a common format for the exchange of beliefs among agents and services in Smart Home. The incorporation of values in the reasoning process presented some advantages such as an autonomy lesser dependent on human input. The flexibility of the architecture is maintained, as the engines of the planning and acting layers can be replaced by any engines that might be needed. This approach has been validated by performing a scenario as the proof of concept, which shows the high timely performance of SMASH agents compared with the literature.

%
%
%
%

\end{document}